# DeepAtrophy: Teaching a Neural Network to Differentiate Progressive Changes from Noise on Longitudinal MRI in Alzheimer's Disease

Mengjin Dong[a], Long Xie[a, b], Sandhitsu R. Das[a, c, d], Jiancong Wang[a], Laura E.M. Wisse[a, b, e], Robin deFlores[c, d], David A. Wolk[c, d], Paul A. Yushkevich[a, b], for the Alzheimer's Disease Neuroimaging Initiative*

[a] Penn Image Computing and Science Laboratory (PICSL), Department of Radiology, University of Pennsylvania, Philadelphia, PA, USA

[b] Department of Radiology, University of Pennsylvania, Philadelphia, Pennsylvania, USA

[c] Department of Neurology, University of Pennsylvania, Philadelphia, Pennsylvania, USA

[d] Penn Memory Center, University of Pennsylvania, Philadelphia, Pennsylvania, USA

[e] Department of Diagnostic Radiology, Lund University, Lund, Sweden

* Data used in preparation of this article were obtained from the Alzheimer's Disease Neuroimaging Initiative (ADNI) database (adni.loni.usc.edu). As such, the investigators within the ADNI contributed to the design and implementation of ADNI and/or provided data but did not participate in analysis or writing of this report. A complete listing of ADNI investigators can be found at: http://adni.loni.usc.edu/wp-content/uploads/how_to_apply/ADNI_Acknowledgement_List.pdf

## Abstract

Volume change measures derived from longitudinal MRI (e.g., hippocampal atrophy) are a well-studied biomarker of disease progression in Alzheimer's disease (AD) and are used in clinical trials to track therapeutic efficacy of disease-modifying treatments. However, longitudinal MRI change measures can be confounded by non-biological factors, such as different degrees of head motion and susceptibility artifact between pairs of MRI scans. We hypothesize that deep learning methods applied directly to pairs of longitudinal MRI scans can be trained to differentiate between biological changes and non-biological factors better than conventional approaches based on deformable image registration. To achieve this, we make a simplifying assumption that biological factors are associated with time (i.e., the hippocampus shrinks over time in the aging population) whereas non-biological factors are independent of time. We then formulate deep learning networks to infer the temporal order of same-subject MRI scans input to the network in arbitrary order; as well as to infer ratios between interscan intervals for two pairs of same-subject MRI scans. When applied to longitudinal MRI scans unseen during training, these networks perform better in the tasks of temporal ordering (89.3%) and interscan interval inference (86.1%) than a state-of-the-art deformation-based-morphometry method ALOHA (76.6% and 76.1% respectively) (Das et al., 2012). Furthermore, the values from the last activation layer of our network can be used to derive a disease progression score that is able to detect a group difference between 58 preclinical AD (i.e., cognitively normal individuals with evidence of beta-amyloid pathology on Positron Emission Tomography (PET)) and 75 beta-amyloid-negative cognitively normal individuals within one year, compared to two years for ALOHA. This suggests that deep learning can be trained to differentiate MRI changes due to biological factors (tissue loss) from changes due to non-biological factors, leading to novel biomarkers that are more sensitive to longitudinal changes at the earliest stages of AD.

# 1. Introduction

Alzheimer's Disease (AD) is characterized by accelerated loss of brain gray matter compared to "normal" aging, particularly in the medial temporal lobe (MTL). In clinical trials of disease-modifying treatments of AD, the measure of hippocampus volume change in the MTL derived from longitudinal MRI is an established biomarker to monitor disease progression and response to treatment. Compared to clinical cognitive tests, MRI-derived biomarkers are more sensitive to change over time, particularly in early stages of AD progression, therefore requiring a smaller cohort and/or shorter trial duration to detect a significant change due to treatment (Ard and Edland, 2011; Jack et al., 2010; Sperling et al., 2011; Weiner et al., 2015).

While there is little debate that longitudinal structural MRI is a critical biomarker for AD clinical trials and disease development estimations (Cullen et al., 2020; Lawrence et al., 2017; Marco Lorenzi et al., 2015), it remains an open question on how to optimally extract measures of change from MRI scans. The straightforward approach of measuring the volume of the hippocampus (or other structure of interest) at multiple time points independently and then comparing them longitudinally suffers from relatively high coefficient of variability in these measurements (Leow et al., 2006; Schuff et al., 2009). Atrophy measures obtained directly from comparing longitudinal MRI scans, e.g., by means of deformable registration, tend to be more sensitive to disease progression, thus reducing several-fold the size of study cohort and/or the duration required in the clinical trials (Fox et al., 2011; Resnick et al., 2003; Weiner et al., 2015).

In recent years, different methods have been developed to estimate atrophy of the hippocampus and other brain structures affected early in AD from longitudinal MRI (Cash et al., 2015; Pegueroles et al., 2017; Xiong et al., 2017). One of the most widely used techniques is deformation-based morphometry (DBM, also known as tensor-based morphometry) (Das et al., 2012; Holland et al., 2095; Hua et al., 2016, 2008; Reuter et al., 2012; Vemuri et al., 2015; Yushkevich et al., 2009), which uses deformable registration to obtain a deformation field mapping locations in the baseline image to corresponding locations in the follow-up image and estimates the change in structures such as the hippocampus by integrating the Jacobian determinant of the deformation field over the hippocampus segmentation in the baseline image (Hua et al., 2012; Lorenzi et al., 2013; Reuter et al., 2010). Another widely used method is the boundary shift integral (BSI) (Gunter et al., 2003; Leung et al., 2010; Prados et al., 2014), in which the displacements of the boundary of a structure of interest, and subsequently, the change in its volume, are inferred by examining the changes in intensity characteristics near the structure's boundary. A number of DBM, BSI, and related longitudinal atrophy estimation techniques were compared on a common dataset by Cash et al. (2015). A challenge in evaluating atrophy techniques is that the ground truth (actual atrophy) is unknown. A common strategy is to examine differences in atrophy rates between individuals at different stages of AD progression, with the hypothesis that a more sensitive method would detect greater differences in the rates of hippocampal atrophy between cohorts with different severity of disease, e.g., clinical AD (greatest atrophy rate), early and late mild cognitive impairment (eMCI and lMCI) and normal controls (NC, most stable) (Cash et al., 2015; Fox et al., 2011). Additionally, same-subject MRI scans taken a short interval of time apart (<2 weeks) are used to evaluate the stability of atrophy estimation methods, since no atrophy is expected to take place over such a short time. The evaluation by Cash et al. (2015) suggests that DBM-style and BSI-style techniques achieve roughly comparable performance for estimating longitudinal atrophy. These techniques remain the state-of-the-art for longitudinal atrophy estimation today.

Neurodegenerative changes in the hippocampus on longitudinal MRI can be obscured by differences in MRI signal that are unrelated to disease progression, such as different amounts of head motion, change in slice plane orientation, susceptibility artifact, and changes in scanner hardware and software. These differences can appear as subtle shifts in the borders of anatomical structures, particularly when these borders are not very strongly defined in the first place. Conventional techniques like DBM and BSI, which rely on image registration and image intensity comparisons to derive atrophy measures, are likely to misinterpret these confounding differences as increases or decreases in hippocampal volume, adding to the overall variance of the measurements. Measurements of atrophy rate in the hippocampus in older adults are expected to be negative (i.e., the volume is reduced over time) (Fox et al., 2011). However, the state-of-the-art DBM pipeline Automatic Longitudinal Hippocampal Atrophy software/package (ALOHA) (Das et al., 2012) reports positive atrophy rates in 26% of beta-amyloid-negative (A-) NC, 23% of beta-amyloid-positive (A+) eMCI (A+ eMCI), and 17% of A+ lMCI longitudinal scan pairs from ADNI (Mueller et al., 2005). We hypothesize that these positive atrophy rate measurements are at least in part caused by registration errors in the DBM pipeline associated with non-biological factors such as motion and MRI artifact, and that deep learning (DL) methods can better account for these factors. The emergence of DL and fast computational power led to a new generation of algorithms that outperform many traditional algorithms in computer vision and medical image analysis (Krizhevsky et al., 2012; Simonyan and Zisserman, 2015). And while there have been a number of DL papers focused on diagnosing AD and predicting future disease progression (summarized in the Discussion), to our knowledge there has been no research using DL to track disease progression from longitudinal MRI. Yet sensitive measures for tracking disease progression, particularly in the earliest stages of the disease, are of critical importance for reducing the cost and duration of clinical trials in AD. In a clinical trial of a disease-modifying treatment for AD, the experimental arm of the trial would be expected to undergo slower rates of disease progression than the placebo arm, and the size and duration of the trial are determined by the ability to detect a statistically significant difference in rates of progression between the trial's arms. If gains attained by adoption of DL in other domains could be extended to the domain of AD disease progression quantification, the potential impact on the cost and duration of AD clinical trials could be substantial.

In this paper, we hypothesize that DL models can be trained to differentiate MRI changes due to biological factors (tissue loss) from changes due to motion, MRI artifacts and other non-biological factors, leading to novel biomarkers that are more sensitive to longitudinal change and can better track disease progression. Since the underlying atrophy rate for each subject is unknown, we cannot directly train a DL model to predict the hippocampal atrophy rates from longitudinal MRI scans. Instead, we make simplifying assumptions about brain atrophy for individuals on the AD continuum, which are used to guide the design of our DL model. Our primary assumption is that atrophy is progressive (i.e., the hippocampus atrophies over time) whereas non-biological factors such as head motion are independent of time. Therefore, a DL model trained to differentiate atrophy from non-biological factors should be able to determine the temporal order of two MRI scans from the same individual. As the first step, we train a DL model to determine the temporal order of same-subject MRI scan pairs that are preprocessed identically and input to the network in arbitrary order. However, it turns out that although it is indeed possible to train a DL model that infers scan order accurately, the values from the output activation layer of such a network are sensitive to *the presence of atrophy* but not to the *magnitude of atrophy*. To address this shortcoming, we make a secondary assumption, that the magnitude of atrophy is approximately proportional to time, with the constant of proportionality varying from individual to individual. To sensitize the DL network to the magnitude of atrophy, we modify it to take two pairs of same-subject MRI scans and train it to infer which of the two pairs has a longer inter-scan interval, in addition to the original task of determining the temporal order of the scans in each pair. The activation values in the output layer of this modified network are expected to track the magnitude of atrophy better than the original network. Furthermore, we expect the network trained in this fashion to infer temporal scan order of pairs of MRI scans, and relative lengths of inter-scan intervals of pairs of scan pairs, better than conventional deformation-based morphometry methods, such as ALOHA. Developing such a DL model and performing this comparison constitutes the first set of experiments in this paper.

Ultimately, our goal is to extract a measure of atrophy from longitudinal MRI scans that is as sensitive as possible to disease progression in early AD. Inspired by recent brain age prediction studies (Cole and Franke, 2017; Liem et al., 2016), which use a mismatch between brain age inferred from imaging data and actual chronological age as a potential biomarker to characterize brain disorders, we employ the mismatch between the inter-scan interval predicted by the DL model and the actual inter-scan interval as a measure of disease progression that accounts for non-biological factors. In our second set of experiments, we compare the ability of our DL-based measures to detect differences in measures of disease progression between groups at different stages of the AD continuum to ALOHA annualized atrophy rate measures. We expect the DL-based measures to detect more significant differences in measures of disease progression between preclinical AD and normal aging than ALOHA. We perform these comparisons using data from the Alzheimer's Disease Neuroimaging Initiative (Jack Jr et al., 2008).

## 2. Methods and Materials

### 2.1 Data Preprocessing

Data used in this study were obtained from the Alzheimer's Disease Neuroimaging Initiative (ADNI, adni.loni.usc.edu). The ADNI was launched in 2003 as a public-private partnership, led by Principal Investigator Michael W. Weiner, MD. The primary goal of ADNI has been to test whether serial magnetic resonance imaging (MRI), positron emission tomography (PET), other biological markers, and clinical and neuropsychological assessment can be combined to measure the progression of mild cognitive impairment and early Alzheimer's disease. For up-to-date information, see www.adni-info.org.

Participants from the ADNI2/GO phases of the ADNI study were included if they had a beta-amyloid PET scan and at least two longitudinal T1-weighted MRI scans with 1x1x1.2 mm$^3$ resolution. 481 participants with 2 to 6 longitudinal T1-weighted MRI scans (0.25 to 5.5 years) were selected (Table 1). Participants were grouped into four cohorts corresponding to progressive stages along the AD continuum: healthy aging (beta-amyloid-negative cognitively normal control, A- NC), preclinical AD (beta-amyloid-positive cognitively normal controls, A+ NC), early prodromal AD (A+ early mild cognitive impairment, A+ eMCI), and late prodromal AD (A+ lMCI). The preclinical AD cohort consists of asymptomatic individuals who are at increased risk of progressing to symptomatic disease, and is of elevated interest for clinical trials of early disease-modifying interventions (Sperling et al., 2014, 2013). A total of 18294 3D longitudinal pairs of MRI scans of bilateral MTL from 29 ADNI sites were used.

Participants were divided a priori into training (n = 155), validation (n = 21) and test (n = 326) subsets. Subjects with only two scans were assigned to the test set since at least three scans per subject are required to train our network, while only one pair of images is required for inference (see below in this section for detailed explanation). In addition, all preclinical AD (A+ NC) subjects were assigned to the test set in order to maximize the number of subjects available to detect group differences between A- NC and A+ NC cohorts. All other subjects were assigned to the training, validation, or test sets with probability of 45%, 5% and 50%, respectively (within each diagnostic group). The validation set was kept small to maximize the number of scans available for inference and used for early stopping of DL network training, rather than for parameter tuning. The age, sex, years of education, and the Mini-Mental State Exam (MMSE) score of subjects the training and test groups are listed in Table 1.

For each scan in each subject, segmentation software ASHS-T1 (Xie et al., 2019) was applied to automatically segment the left and right medial temporal lobe (MTL) subregions, including the hippocampi. As a preprocessing step in ASHS-T1, MRI scans were upsampled to 1x0.5x0.6mm$^3$ resolution using a non-local mean super-resolution technique (Coupé et al., 2013; Manjón et al., 2010). The ASHS-T1 segmentation was then used to crop out a ~8.5x6.0x6.5cm$^3$ area from the upsampled image, centered on the MTL on each side of the brain. For each pair of longitudinal MRI scans, rigid registration was performed using ANTs (Avants et al., 2007) between the cropped MTL regions using the normalized cross-correlation metric. To ensure that both scans in a pair are preprocessed identically, the 6-parameter transformation matrix was factored into two equal matrices, and both scans were resampled into a common half-way space by applying the corresponding matrix (Yushkevich et al., 2009). To further avoid the possibility of bias due to preprocessing, registrations were conducted twice with each one of the two images in the pair being input once as the "fixed" image and once as the "moving" image. Thus, for each image pair in their original space, two pairs of rigidly aligned images are created. This two-way symmetric registration process ensures the subsequent experiments undergo exactly the same preprocessing and interpolation operation regardless of the temporal order of the images. The total number of pairwise rigid registrations (19708) was too large to manually check for registration errors. Instead, we computed the Structural SIMimilarity (SSIM) metric (Wang et al., 2004) for each pair of registered scans, and rejected pairs with SSIM < 0.6 to guarantee high image quality (e.g. no ringing effect) and alignment. This resulted in 1414 scan pairs (7.2%) being rejected.

The intensity of both the training and test images is normalized to the unit normal distribution. All images are randomly cropped again to a fixed size (48x80x64 voxels) around the MTL region (segmented by ASHS-T1) to maximally preserve the ROI before inputting to the network. Data augmentation is only applied to the training set, and it includes random flip with 50% probability on one of the three dimensions, and thin plate spline (TPS) transformation with 10 randomly selected points. All data augmentation is applied in the same way for two images in an image pair.

Table 1. Characteristics of the selected ADNI2/GO participants whose T1 MRI scans were used for the DeepAtrophy and ALOHA experiments. All subjects in the training and test set had 2 to 6 scans between 0.25 and 5.5 years from the baseline. Abbreviations: n = number of subjects; A+/A- = beta-amyloid positive/negative; NC = cognitively normal adults; eMCI = early mild cognitive impairment; lMCI = late mild cognitive impair; Edu = years of education; MMSE = mini-mental state examination; ALOHA = Automatic Longitudinal Hippocampal Atrophy software/package; F = female; M = male.

|  | Training Set (n = 155) | | | | Test Set (n = 326) | | | |
|---|---|---|---|---|---|---|---|---|
|  | A- NC (n = 64) | A+ NC (n = 0) | A+ eMCI (n = 49) | A+ lMCI (n = 42) | A- NC (n = 103) | A+ NC (n = 85) | A+ eMCI (n = 79) | A+ lMCI (n = 59) |
| Age (years) | 71.5 (5.7) | - | 73.3 (7.4) | 71.9 (7.2) | 72.4 (6.3) | 75.2 (5.8)** | 74.0 (6.9) | 72.5 (6.5) |
| Sex | 27F 37M | - | 16F 33M | 21F 21M | 56F 47M | 58F 27M | 39F 40M | 28F 31M |
| Edu (years) | 17.4 (2.3) | - | 16.2 (3.0)* | 16.8 (2.7) | 16.8 (2.4) | 16.0 (2.7)* | 15.3 (2.9)*** | 16.4 (2.7) |
| MMSE | 29.1 (1.2) | - | 28.0 (1.5)**** | 26.7 (1.8)**** | 29.1 (1.2) | 29.0 (1.1) | 28.1 (1.6)**** | 27.3 (1.9)**** |

*Notes*: *, $p < 0.05$; **, $p < 0.01$; ***, $p < 0.001$; ****, $p < 0.0001$. All statistics in the train and test set are in comparison to the corresponding A- NC group. *Standard deviation* is reported in parenthesis. Independent two-sample *t*-test (continuous variables with normal distribution, for age and education), Mann–Whitney *U* test (continuous variable with non-normal distribution, for MMSE) and contingency $\chi^2$ test (for sex) were performed.

Figure 1. Diagram of the DeepAtrophy deep learning algorithm for quantifying progressive change in longitudinal MRI scans. During training, DeepAtrophy consists of two copies of the same basic sub-network with shared weights, whose outputs feed into a shared fully connected layer. This basic sub-network takes as input two MRI scans from the same individual input in arbitrary temporal order. The basic sub-network consists of a 3D ResNet image classification network of 50 layers (Chen et al., 2019; He et al., 2015). The overall network takes as input three or four same subject-images, in arbitrary order, and with constraint that the scan pair input into one of the basic sub-networks has a time interval that includes the time interval for the scan pair input into the other. DeepAtrophy minimizes a weighted sum of two loss functions: the scan temporal order (STO) loss, which measures the ability of the basic network to correctly infer the temporal order of the two input scans; and the relative interscan interval (RISI) loss, which measures the ability of the complete network to determine which of the input scan pairs has a longer interscan interval. During testing, the basic network is applied to pairs of same-subject scans, and its output is used to derive a single measure of disease progression, the predicted interscan interval (PII).

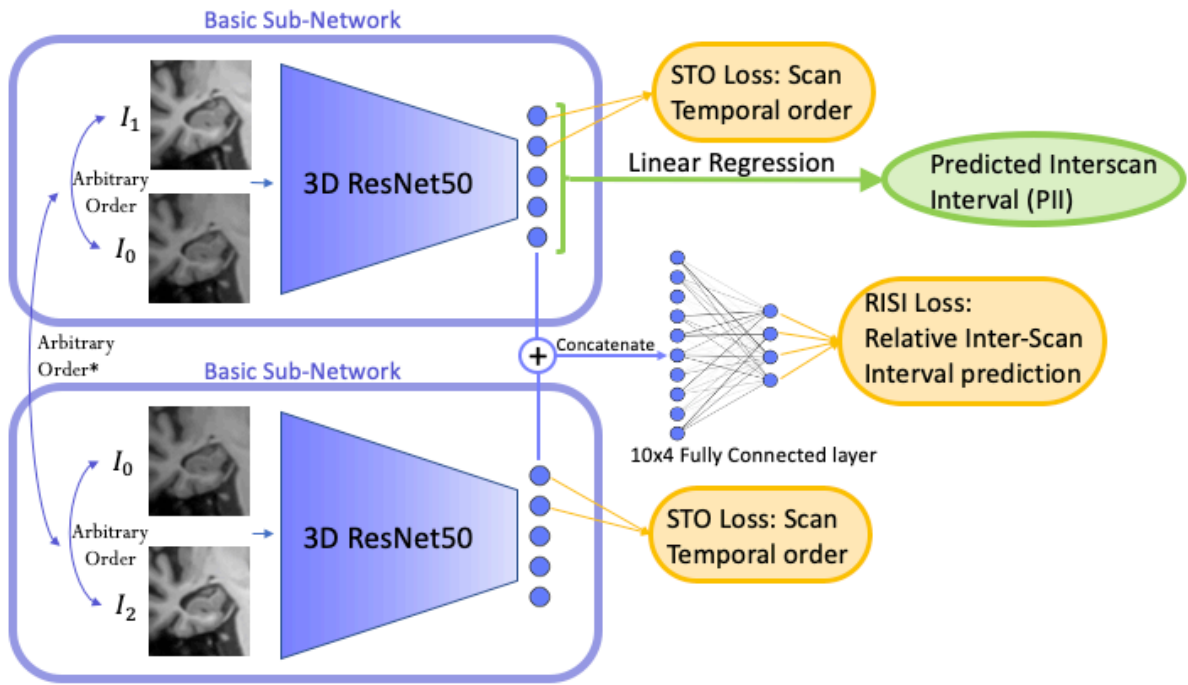

## 2.2 Basic Network Structure and the Scan Temporal Order (STO) Loss

The basic building block of our DL algorithm is a deep convolutional neural network that takes two longitudinal MRI scans of the same subject as inputs and classifies them as being in correct temporal order (scan $I_1$ has later date than scan $I_0$) or reverse temporal order (scan $I_1$ has earlier date than scan $I_0$). As argued in the Introduction Section, the network's ability to detect the temporal order of arbitrarily ordered pairs of scans would suggest that the network is sensitive to biological changes that are progressive in time, as opposed to non-biological factors that are independent of time. The underlying network is the ResNet50 deep residual learning network (He et al., 2015) used extensively for image classification in computer vision. In our implementation, we used a 3D version of ResNet50 pretrained on medical images of multiple organs (not including ADNI data) named MedicalNet (Chen et al., 2019). We empirically chose the 50 layers pretrained model as the backbone structure of our network to avoid under- and over-fitting. After the last average pooling layer of the original ResNet50 model, a fully connected (FC) layer with a five-dimensional output vector was applied. The first two of these five outputs are input to a cross-entropy loss in order to infer the scan temporal order. We refer to this loss as the *Scan Temporal Order (STO) loss*. All five outputs are used for the *relative inter-scan interval (RISI) loss*, defined in the next section.

## 2.3 Inference of Relative Inter-Scan Interval (RISI)

In our initial experiments, the basic network structure with the STO loss was able to detect the presence and directionality of atrophy between two scans but the activation values in the last layer of the network showed poor sensitivity to the magnitude of atrophy. For example, the network would generate very similar activation values when applied to scans with a one-year interscan interval and a five-year interscan interval. This made it difficult to derive a suitable disease progression measure from the network outputs. In contrast, ALOHA, a conventional deformation-based morphometry approach, generates atrophy measures that on average scale linearly with the inter-scan interval.

In order to sensitize network outputs to the magnitude of atrophy, we adopt a Siamese network architecture (Bertinetto et al., 2016) shown in Figure 1. The Siamese network consists of two copies of the basic network outlined above, with shared weights. The input to the Siamese network consists of two pairs of scans of the same individual, with one arbitrarily chosen pair having a longer interscan interval than the other. Furthermore, the scans are selected in such a way that the interscan interval of one pair is strictly within the interscan interval of the other pair (e.g., scan pairs {0 year, 1 year} and {0 year, 5 year} are allowed, but scan pairs {0 year, 1 year} and {2 year, 5 year} are not), ensuring that the atrophy in the pair with the longer interscan interval is greater than in the pair with the shorter inter-scan interval. The goal of the Siamese network is to sensitize the basic network to the magnitude of atrophy between its inputs by having it produce features in its output layer that scale with the interscan interval.

During training, given the four inputs to the Siamese network, we compute the ratio of the absolute value of the interscan interval for the first pair and the absolute value of the interscan interval for the second pair, and assign it to four categories: $[0,\ 0.5)$, $[0.5,\ 1)$, $[1,\ 2)$, $[2, +\infty)$. The Siamese network is then trained to correctly infer one of these categories from its inputs. The final layers of the two copies of the basic network, each consisting of five outputs, are concatenated and input into a 10x4 FC layer. The four-element output of this layer is input to the cross-entropy loss, which we call the *Relative Inter-Scan Interval (RISI) loss.*

Formulating the problem of interscan interval ratio inference as a classification problem with the four categories above, as opposed to a regression problem, is driven by two considerations. On the one hand, we empirically found the networks with the categorical loss easier to train. On the other hand, the categorical loss encodes the progressive relationship between interscan interval and the output of the basic sub-network (presumably a measure of atrophy under our assumptions) without explicitly assuming this relationship to be linear, which is likely not the case for most individuals.

The overall network, which we call *DeepAtrophy,* is trained by minimizing the weighted sum of the STO losses for the two Siamese sub-networks and the RISI loss for the overall network. Higher weighting of the RISI losses encourages the network to focus more effort on detecting the magnitude of disease progression, while higher weighting of the STO loss encourages it to focus more effort on detecting the presence/direction of progression. To avoid any possible bias related to preprocessing when training the Siamese network with the RISI loss, we randomly choose for each scan pair the preprocessing result where the first image in the pair was used as the fixed image during registration or the preprocessing result where the second image was the fixed image.

Lastly, we emphasize that two scan pairs per subject are required when training the DeepAtrophy network, but when the network is applied at test time, only the basic sub-network is used, and one scan pair per subject is required, as described below.

## 2.4 Predicted-to-Actual Interscan Interval Ratio (PAIIR)

DBM and BSI methods yield an intuitive quantitative measure such as annualized loss of hippocampal volume that can serve as disease progression and treatment response biomarkers for clinical trials. In this section, we devise a similar biologically meaningful measure for our deep learning-based approach, which does not explicitly measure atrophy in any particular anatomical structure. We follow the example of recent brain-age prediction studies (Cole and Franke, 2017; Liem et al., 2016), in which the mismatch between a person's actual age and their "brain age" predicted from neuroimaging and/or other biomarkers is used to characterize individuals in terms of resilience vs. vulnerability to the aging process. Adopting this idea, we propose to use the mismatch between the actual interscan interval between two longitudinal MRI scans and the interscan interval predicted from these scans by deep learning as a measure of disease progression. This mismatch is expected to be greater in patients with more advanced AD, and smaller in healthy aging. In other words, for a person with advanced disease, two scans would appear to the network to be farther apart in time than they really are, while for a person with little or no disease, the scans would appear to be closer in time than they really are. We call this measurement the "Predicted-to-Actual Interscan Interval Ratio" (PAIIR).

PAIIR is defined as follows. After training the DeepAtrophy network, we apply the basic sub-network to pairs of same-subject scans $I_{t_1}^k, I_{t_2}^k$ and extract the five activation scores from its last layer, $\left(x_1^k, x_2^k, ..., x_5^k\right) = \left(x_1^k\left(I_{t_1}^k, I_{t_2}^k\right), x_2^k\left(I_{t_1}^k, I_{t_2}^k\right), ..., x_5^k\left(I_{t_1}^k, I_{t_2}^k\right)\right)$. These five scores are the output of the basic sub-network and are expected to reflect both the presence and magnitude of disease progression between $I_{t_1}^k$ and $I_{t_2}^k$. We model the relationship between these scores and the signed interscan interval $t_2^k - t_1^k$ as a linear relationship at the group level, i.e.,

$$t_2^k - t_1^k = \beta_0 + \beta_1 x_1^k + \beta_2 x_2^k + \beta_3 x_3^k + \beta_4 x_4^k + \beta_5 x_5^k + \varepsilon \qquad (1)$$

Note that $I_{t_1}^k$ and $I_{t_2}^k$ are input to the network in random order and that $t_2^k - t_1^k$ can be positive or negative. We learn the parameters $\beta_0, \beta_1, ..., \beta_5$ by least square fitting 2250 scan pairs from A-NC cohort in the training data. At test time, we apply the basic sub-network to each pair of longitudinal MRI scans, and derive the Predicted Interscan Interval (PII) as

$$PII\left(I_{t_1}^k, I_{t_2}^k\right) = \beta_0 + \beta_1 x_1^k + \beta_2 x_2^k + \beta_3 x_3^k + \beta_4 x_4^k + \beta_5 x_5^k \qquad (2)$$

The PAIIR was calculated as

$$PAIIR(I_{t_1}^k, I_{t_2}^k) = \frac{PII(I_{t_1}^k, I_{t_2}^k)}{t_2^k - t_1^k} \quad (3)$$

PAIIR is an indicator of the rate of disease progression and is envisioned as a surrogate to the traditional atrophy rate measurement in DBM. Absolute values of PAIIR larger than one are suggestive of disease progression occurring faster than what is expected for the A- NC group. Similar to atrophy rate measures derived from DBM and BSI methods, we expect PAIIR to be greater on average in patients in more advanced stages of AD.

## 2.5 Statistical Tests

In the first set of experiments, we compared the accuracy of temporal ordering of scan pairs (explicitly maximized by the STO loss) and the accuracy of longer vs. shorter interscan interval detection for pair of scan pairs (explicitly maximized by the RISI loss) between DeepAtrophy and ALOHA. For brevity, we refer to these measures as "STO accuracy" and "RISI accuracy". Accuracy was measured on the held-out test data, with DeepAtrophy and ALOHA applied to the same set of pairs. STO accuracy for DeepAtrophy was computed by comparing the predicted class in the STO loss to the actual scan ordering. STO accuracy for ALOHA was computed by comparing the sign of the annualized hippocampal volume change measure to the scan temporal ordering (i.e., expecting ALOHA to report negative atrophy for a pair of scans in correct temporal order). RISI accuracy for DeepAtrophy was calculated by taking the predicted and the actual interscan intervals of two pairs of scans and determining if the scan pair with the larger predicted interscan interval also has the larger actual interscan interval. For ALOHA, the RISI accuracy was calculated by measuring total hippocampal volume change for each scan pair and determining whether the scan pair with the larger absolute value of volume change had a longer interscan interval. STO accuracy for DeepAtrophy and ALOHA is reported as the area under the receiver operating characteristic curve (AUC). To test for the significance in the difference between AUCs of the two methods, DeLong's test was performed with R package "pROC" (Robin et al., 2011).

We also analyzed the utility of the PAIIR measure from DeepAtrophy for detecting group differences in rates of disease progression between cohorts at different stages of the AD continuum. We compared effect sizes for group comparisons between A+ NC, A+ eMCI and A+ lMCI groups and the A- NC group respectively obtained using PAIIR to the corresponding effect sizes obtained using ALOHA annualized hippocampal volume change measures. In addition, we compared our longitudinal measurements with longitudinal Preclinical Alzheimer's Cognitive Composite (PACC) score, a standard cognitive test crafted specifically for detecting subtle changes in pre-symptomatic disease (Donohue et al., 2014). These comparative analyses were carried out over time intervals of different duration (180 to 400 days; 400 to 800 days) to "simulate" different clinical trial scenarios. The test dataset was reduced to 256 subjects and 246 subjects in total for the 180-to-400-day and 400-to-800-day intervals, respectively (see Supplementary S1 for detailed characteristics). When subjects had more than two scans (or PACC scores) available in this time window, we computed summary scores as follows. For ALOHA, we used the baseline hippocampal volume from ASHS-T1 and pairwise volume change measures between the baseline image and each follow-up image to estimate the hippocampal volume at each time point and fitted a linear model to these measurements. The slope of the linear fit was taken as the summary atrophy measure. For DeepAtrophy, we followed a similar approach, using PII instead of volume change, and using zero for the baseline measurement. For PACC, we also followed this linear fitting approach, however, most subjects had only two tests within the 400-day interval. Additionally, all summary scores (DeepAtrophy, ALOHA, PACC) were corrected for age based on a linear model. For each of the above approaches, we conducted Student's t-test (unpaired, two-sided) between the corresponding measure of disease progression in each disease group (A+ NC, A+ eMCI, and A+ lMCI) and the A-NC group.

Additionally, for DeepAtrophy and ALOHA, we estimated the minimum sample size required to detect a 25%/year and 50%/year reduction in the atrophy rate of each disease stages (A+ NC, A+ eMCI, and A+ lMCI) relative to the to the mean atrophy rate of the A- NC group in a hypothetical clinical trial. This calculation envisions a clinician trial in which participants are patients at a given disease stage (e.g., preclinical AD) and the intervention successfully slows disease progression by 25% or 50% relative to "normal" brain atrophy in this age group (Fox et al., 2011). The sample size describes the minimal number of participants in the treatment and placebo arms of the clinical trial needed to detect a significant difference between the two arms with a two-sided significance level α = 0.05 and power 1-β = 0.8. The sample size is calculated as

$$N = \left[ \frac{(z_{1-\alpha/2} + z_\beta) S_{PAT}}{0.25 * (\bar{A}_{PAT} - \bar{A}_{CTL})} \right]^2$$

where $\bar{A}_{PAT}$ and $\bar{A}_{CTL}$ are the sample means of the patient and control group, and $S_{PAT}$ is the sample standard deviation of the patient group. The 95% confidence interval for each sample size measurement was computed with the bootstrap method (B. Efron, 1979).

## 2.6 Implementation Details

Experiments were conducted on a Titan 2080 Ti GPU with 8 GB memory. Images are cropped to 48x80x64 to the network. 71656 combinations of image pairs (each input consists of two image pairs of the same subject) were trained in the DeepAtrophy with a learning rate of 0.001, batch size of 15, with 15 epochs for 15 hours. In the following report of the results, we adopt an optimal set of weights of the STO loss and RISI loss to be 1 and 1 respectively. In the Supplementary S3 we compare different weighting schemes of the STO and RISI losses.

# 3. Results

## 3.1 Scan Temporal Order (STO) Inference Accuracy

Table 2 compares the mean accuracy of detecting the correct temporal order of a single pair of same-subject scans (STO accuracy) from the held-out test subset input in random order to DeepAtrophy and ALOHA algorithms. The same set of 4569 input pairs was used to evaluate both algorithms, selected at random from among all available scan pairs, with no maximum cutoff for interscan interval. STO accuracy for DeepAtrophy was 89.3% of all scan pairs, compared to 76.6% for ALOHA. For both methods, STO accuracy was lower for less impaired groups, as would be expected since there is less underlying biological change for the same time interval than in more impaired groups. The receiver operating characteristic (ROC) plot in Figure 2 further contrasts the ability of DeepAtrophy and ALOHA in inferring scan temporal order. For all groups, the area under the curve (AUC) for DeepAtrophy is significantly higher than for ALOHA (p-value < 2.2e-16).

Table 2. Accuracy of our DeepAtrophy and hippocampus atrophy extracted using ALOHA (Das et al., 2012), a conventional registration-based method, in inferring the scan temporal order (STO) of same-subject scan pairs input in arbitrary order (STO accuracy). For ALOHA, we consider it to be "correct" if the sign of hippocampal atrophy is negative for scans input in chronological order, and positive for scans in reverse chronological order. Accuracy is expected to be lower for less impaired groups because there is less underlying biological change for the same time interval than in more impaired groups. Abbreviations: ALOHA = Automatic Longitudinal Hippocampal Atrophy software/package; A+/A- = beta-amyloid positive/negative; NC = cognitively normal adults; eMCI = early mild cognitive impairment; lMCI = late mild cognitive impair.

|  | A- NC | A+ NC | A+ eMCI | A+ lMCI | Average |
|---|---|---|---|---|---|
| ALOHA | 73.6% | 74.5% | 77.3% | 82.9% | 76.6% |
| DeepAtrophy | 87.6% | 87.7% | 89.8% | 93.2% | 89.3% |

Figure 2. Area under the receiver operating characteristic (ROC) curve (AUC) for the scan temporal order (STO) inference experiments using DeepAtrophy and ALOHA in the held-out test subset. Greater AUC for DeepAtrophy indicates greater accuracy in inferring the temporal order of scans, which we hypothesize is indicative of greater ability to separate biological changes in longitudinal MRI from nonsystematic changes such as subject motion. Abbreviations: ALOHA = Automatic Longitudinal Hippocampal Atrophy software/package; MCI = mild cognitive impairment.

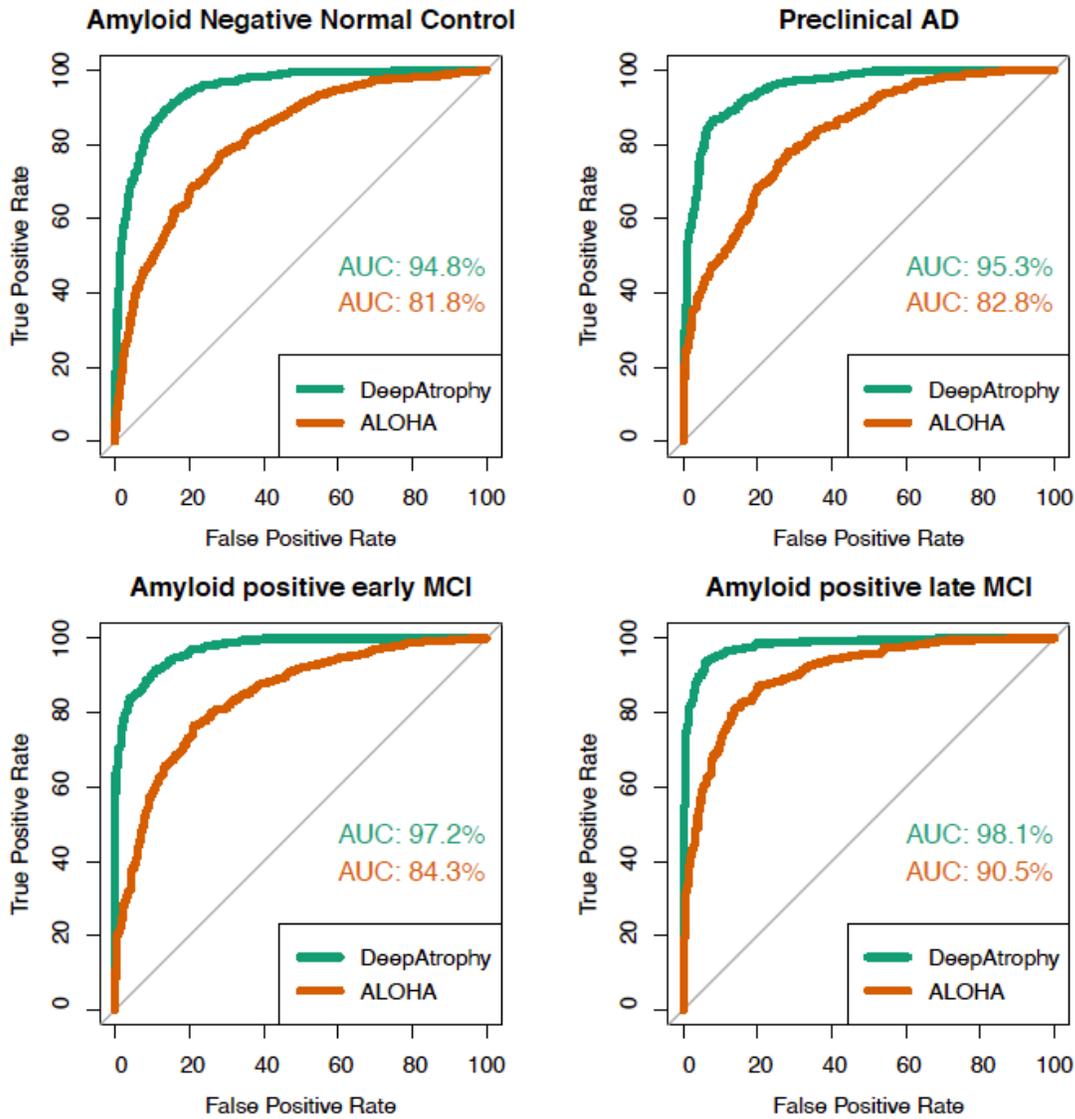

## 3.2 Relative Inter-Scan Interval (RISI) Inference Accuracy

Table 3 compares the mean accuracy of DeepAtrophy and ALOHA in the task of inferring which out of two pairs of same-subject scans has a longer interscan interval (RISI accuracy). This evaluation used data from all subjects in the held-out test set who had at least three scans, with no maximum cutoff for the interscan interval. The two approaches were applied to the same set of input scans. DeepAtrophy has higher RISI accuracy (86.1%) compared with ALOHA (76.1%). This suggests that deep learning can infer not only the presence, but also the magnitude of disease progression from a pair of MRI scans.

Table 3. Comparison of RISI accuracy for DeepAtrophy and ALOHA. Given two pairs of scans from the same subject of different interscan-intervals, a method sensitive to underlying biological change should be able to correctly detect which scan pair has a longer inter-scan interval. Abbreviations: ALOHA = Automatic Longitudinal Hippocampal Atrophy software/package; A+/A- = beta-amyloid positive/negative; NC = cognitively normal adults; eMCI = early mild cognitive impairment; lMCI = late mild cognitive impair.

|  | A- NC | A+ NC | A+ eMCI | A+ lMCI | Average |
|---|---|---|---|---|---|
| ALOHA | 70.3% | 72.7% | 79.5% | 83.4% | 76.1% |
| DeepAtrophy | 82.1% | 86.5% | 87.4% | 89.0% | 86.1% |

## 3.3 Visualizing Disease Progression in Individual Subjects

Figure 3 uses spaghetti plots to visualize the individual trajectories of subjects in the held-out test set measured using DeepAtrophy, ALOHA and PACC approaches. In these plots, we only include data between baseline and 400 days after baseline. For each subject and each method, the plot shows the corresponding measure (predicted interscan interval for DeepAtrophy, hippocampal volume change for ALOHA, score difference for PACC) between baseline and each follow-up timepoint available for that measure. For DeepAtrophy, the progression measure should increase with time (since the predicted interscan interval is expected to be consistent with the actual interscan interval), whereas for ALOHA and PACC, the progression measure (hippocampal volume, PACC score) is expected to decrease with time. Moreover, we expect the relationship between time interval and each progression measure to be approximately linear, in aggregate. Even if within each individual disease progression does not follow a linear trajectory, given that individuals are observed at different times in the disease course, on average we expect to observe a close to linear relationship. Indeed, for ALOHA, the conventional method, we observe a close to linear relationship overall, although there is a great deal of variation among the individual trajectories. Reflecting the relatively low STO accuracy of ALOHA, a number of trajectories are in the upper quadrant of the coordinate space (i.e., hippocampal volume increase). In contrast, the trajectories of DeepAtrophy are almost entirely in the upper right quartile (consistent with its high STO accuracy), but the relationship between the predicted interscan interval and time is sublinear, suggesting that DeepAtrophy is sensitized to short-term longitudinal changes to a greater extent than to longer-term changes. Notably, when the weight of the RISI loss in DeepAtrophy is reduced (shown in spaghetti plots in Supplemental Figure S1), the relationship becomes even more sublinear, highlighting the importance of the RISI loss in detecting not just the presence of a systematic change between longitudinal scans, but also its magnitude. The trajectories for PACC are much noisier than that of the MRI-based measures. Since most subjects had the PACC test only once a year, the linearity of PACC cannot be observed in this plot. For all three measurements, individuals with more severe disease tend to have trajectories with a higher slope than healthier individuals, suggesting that all three measurements can differentiate differences in the rates of disease progression across the spectrum of AD.

Figure 3. Comparison of (a) DeepAtrophy predicted interscan interval (PII), (b) Automatic Longitudinal Hippocampal Atrophy software/package (ALOHA) volume change, and (c) Preclinical Alzheimer's Cognitive Composite (PACC) score change for individual subjects within 400 days from baseline. For DeepAtrophy, the predicted interscan interval, as an indicator of brain change, is expected to be above zero. For ALOHA and PACC, the volume/score change is expected to be below zero to represent brain atrophy/cognitive decline. Abbreviations: A+/A- = beta-amyloid positive/negative; NC = cognitively normal adults; eMCI = early mild cognitive impairment; lMCI = late mild cognitive impair.

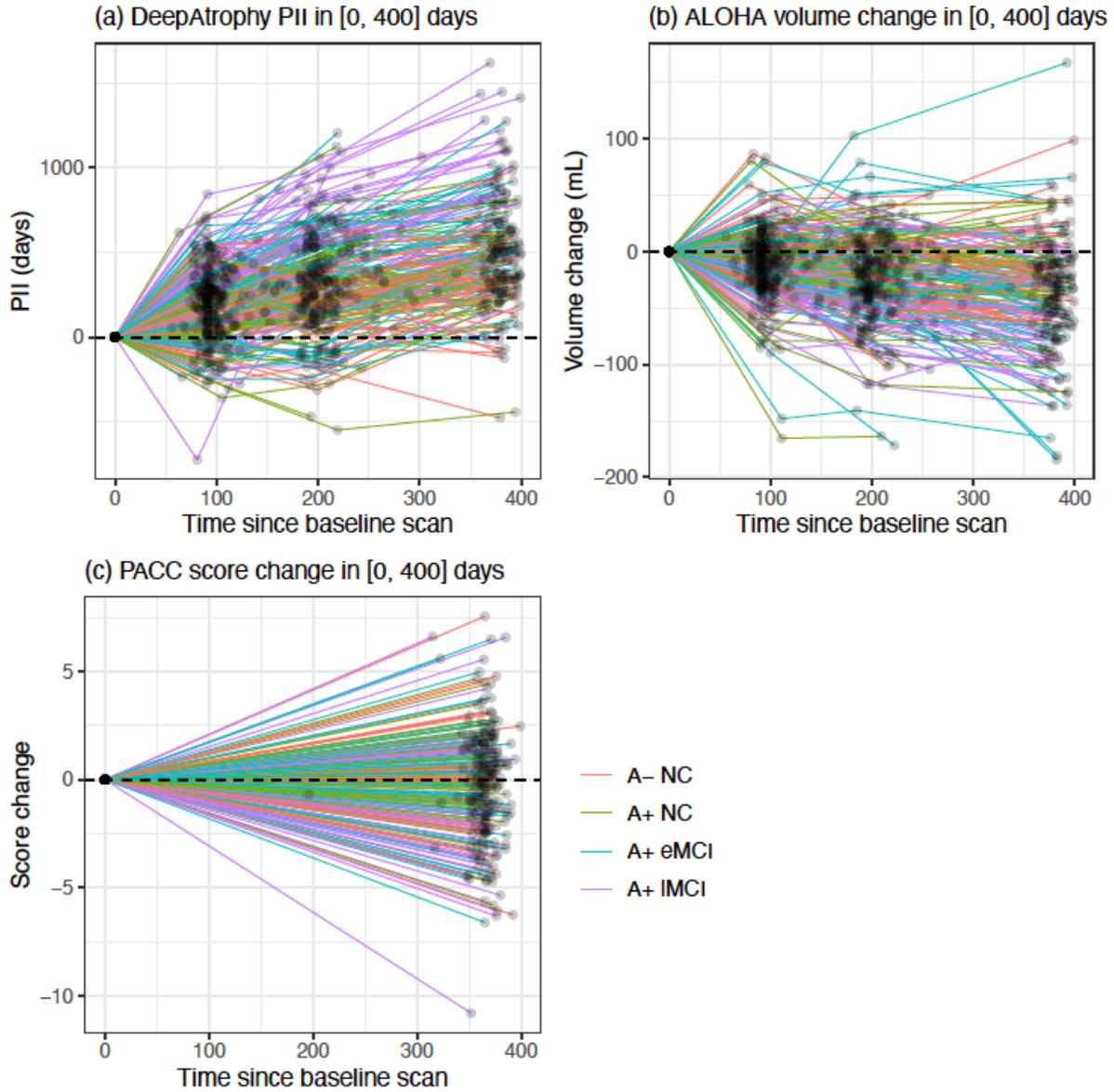

### 3.4 Group Differences in Rates of Disease Progression

In the remaining experiments, we compare the measures of disease progression generated by DeepAtrophy, ALOHA and PACC between groups of ADNI participants at different stages of the AD continuum. Each of the three "disease" groups: preclinical AD (A+ NC), early prodromal AD (A+ eMCI) and late prodromal AD (A+ lMCI) is compared to the "control" group (A- NC) using data from the held-out test set. For each subject, the baseline timepoint and all available follow-up timepoints between 180 and 400 days after baseline were used to compute a single individual disease progression score for DeepAtrophy (PAIIR), ALOHA (annualized hippocampal atrophy rate), and PACC (annualized change in score), as described in Section 2.5. The 180/400-day cutoff for follow-up images was chosen to simulate a clinical trial with limited duration. The same set of subjects are used in the three experiments, and the same scans are used in the DeepAtrophy and ALOHA experiments. Additional experiments were performed using follow-up timepoints between 400 and 800 days after baseline and are mainly presented in the Supplementary Material S4.

Figure 4 compares the ability of DeepAtrophy, ALOHA, and PACC in detecting differences in the rates of disease progression between each disease group and the control group. For all three measures, the average rates of progression increase with the severity of disease. However, notably, only DeepAtrophy detects a statistically significant difference between preclinical AD and the control group (p=0.028), suggesting that DeepAtrophy may be a better biomarker to detect disease progression at this early disease stage. For early and late prodromal AD, both DeepAtrophy and ALOHA detect statistically significant differences relative to the control group (p-value < 0.001), with p-values smaller in absolute terms for DeepAtrophy than ALOHA. Unlike the MRI-based measures, with PACC, only the difference between A+ lMCI and controls can be detected using the 400-day cutoff.

The corresponding plot for the 400 to 800 days cutoff in Supplemental Figure S2 shows consistent results. For this longer cutoff, both ALOHA and DeepAtrophy detect a statistically significant difference between the preclinical AD and control groups, but the p-value is smaller for DeepAtrophy in absolute terms.

Figure 4. Comparison of the ability of DeepAtrophy PAIIR measure, ALOHA atrophy rate, and PACC score change rate to detect differences in the rates of disease progression between normal controls (A- NC) and three disease groups: preclinical AD (A+ NC), early prodromal AD (A+ eMCI) and late prodromal AD (A+ lMCI) using follow-up timepoints between 180 to 400 days from baseline. In each subplot, independent samples t-tests were conducted, after correcting for age, in comparison with the control group and the p-values were shown for each comparison.

Notably, DeepAtrophy is the only measurement significantly differentiating preclinical AD (A+ NC) from A- NC and yields higher level of significance in differentiating A+ eMCI and A+ lMCI from A- NC compared to ALOHA and PACC, suggesting that DeepAtrophy may be a better biomarker to detect disease progression at an early disease stage. Abbreviations: PAIIR = predicted-to-actual interscan interval ratio; ALOHA = Automatic Longitudinal Hippocampal Atrophy software/package; PACC = Preclinical Alzheimer's Cognitive Composite; A+/A- = beta-amyloid positive/negative; NC = cognitively normal adults; eMCI = early mild cognitive impairment; lMCI = late mild cognitive impair; N = number of subjects in the diagnosis group.

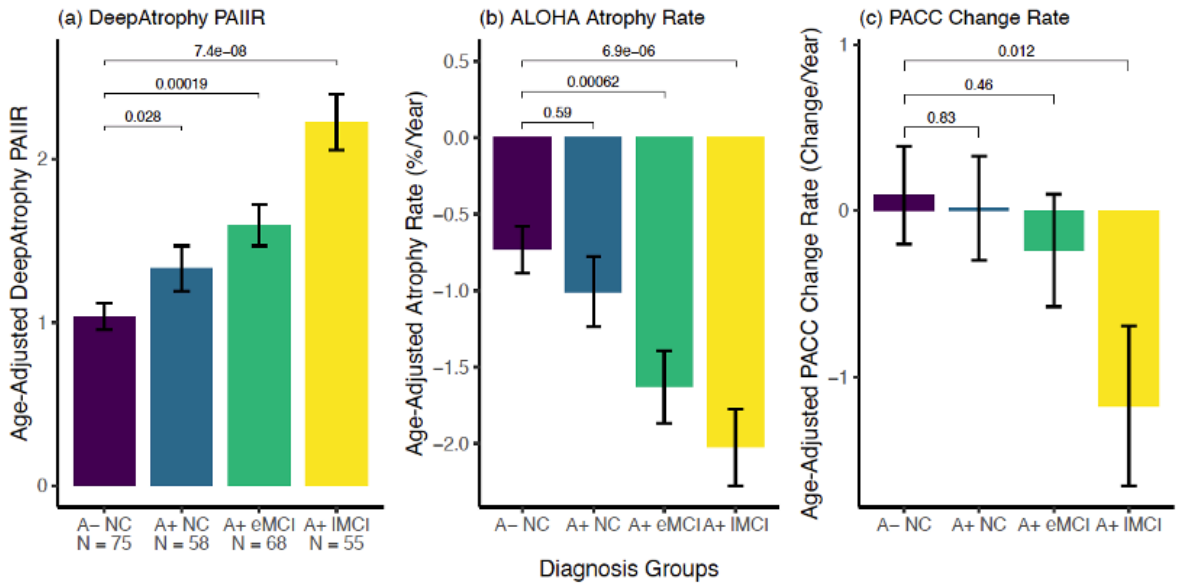

## 3.5 Sample Size Estimation for a Hypothetical Clinical Trial

Table 4 presents sample size estimates (and 95% confidence intervals) for different hypothetical clinical trial scenarios using either DeepAtrophy or ALOHA to track disease progression. Different clinical trial scenarios include participants with different severity of AD (preclinical AD, early prodromal AD, late prodromal AD), different duration (1 vs. 2 years), and different expected reduction in rate of disease progression (25% vs. 50%) in trial participants receiving treatment. As described in Section 2.5, the reduction is computed relative to the rate of progression in controls. In all scenarios, the sample size calculation is based on the statistics (mean and variance) of the four diagnosis groups plotted in Figure 4 (and Supplemental Figure S2 for the 2-year trial). For all scenarios, DeepAtrophy is associated with a smaller sample size estimate (in absolute terms) than ALOHA, although 95% confidence intervals overlap. The results are particularly encouraging for preclinical AD scenarios, where there has been very little prior literature reporting sample size estimates. With a two-year trial and expected reduction of 50%, DeepAtrophy required a trial with the sample size of 374, compared with 582 for ALOHA.

Table 4. Sample size estimates (and 95% confidence interval in bracket) to power a one-year or two-year clinical trial to detect a 25%/year or 50%/year reduction (relative to A- NC) in the rate of disease progression of each patient group. See section 2.5 for sample size calculation. Measurements with a smaller sample size estimation was highlighted in bold. Abbreviations: ALOHA= Automatic Longitudinal Hippocampal Atrophy software/package; A+/A- = beta-amyloid positive/negative; NC = cognitively normal adults; eMCI = early mild cognitive impairment; lMCI = late mild cognitive impair.

| | | 1-year trial (180 to 400 days) | | 2-year trial (400 to 800 days) | |
| | | 25% reduction | 50% reduction | 25% reduction | 50% reduction |
|---|---|---|---|---|---|
| A+ NC | ALOHA | 6832 [1253, 1788342] | 1708 [323, 321167] | 2328 [636, 120045] | 582 [155, 26301] |
| | DeepAtrophy | **4492** [733, 1023775] | **1123** [180, 292956] | **1499** [403, 28998] | **374** [101, 6103] |
| A+ eMCI | ALOHA | 1289 [433, 7825] | 322 [110, 1970] | 800 [294, 3614] | 200 [72, 915] |
| | DeepAtrophy | **907** [345, 3869] | **226** [88, 897] | **260** [135, 501] | **65** [34, 125] |
| A+ lMCI | ALOHA | 455 [209, 1195] | 113 [52, 299] | 328 [130, 949] | 82 [32, 237] |
| | DeepAtrophy | **253** [136, 503] | **63** [33, 127] | **145** [67, 300] | **36** [17, 73] |

# 4. Discussion

In this paper, we hypothesized that multiple non-biological or non-systematic factors such as imaging artifacts and subject motion contribute to the variance in atrophy estimates derived using conventional deformation-based methods from longitudinal MRI; and that it is possible to engineer deep learning methods that are more robust to these factors. The novelty of our approach lies in formulating the problem of measuring longitudinal disease progression in the MTL region in terms of implicitly decomposing longitudinal MRI image pairs into systematic/progressive components (those that track with time) and non-systematic/random components (which are independent of time). By assuming that most older adults' brain changes are progressive, we reframe the problem of detecting disease progression as the problem of inferring temporal information (scan temporal order and relative inter-scan interval) from pairs of longitudinal MRI scans. By using relative measures of time when training the neural network, rather than absolute ones (i.e., inferring actual interscan interval from a pair of scans), our approach implicitly accounts for different rates of disease progression in different individuals. Overall, our results show a deep learning network (DeepAtrophy) explicitly trained to infer temporal information from scan pairs performs this task on test data much more accurately that ALOHA hippocampal atrophy measures. Moreover, activation values from the last layer of DeepAtrophy can be combined to a single scalar measure of disease progression (PAIIR) that performs on par with ALOHA hippocampal atrophy at discriminating patient groups along the AD continuum, indeed achieving greater effect sizes in absolute terms in most of our experiments.

## 4.1 Deep Learning for AD Longitudinal Biomarkers

Current deep learning techniques for AD analysis are focused mainly on the diagnosis and prediction of structural change or cognitive scores of AD (Li and Fan, 2019; Parisot et al., 2018; Spasov et al., 2019; Zhang et al., 2017). It includes classification of the future AD stages (Basu et al., 2019) or time of conversion from one state to another (Lee et al., 2019; Lorenzi et al., 2019), and regression of biomarker values, such as cognitive scores and ventricle volumes (Ghazi et al., 2019; Jung et al., 2019). Prediction of AD stage and conversion time were mainly conducted with Recurrent Neural Networks (RNN), including Long-Short Term Memory (LSTM) networks (Ghazi et al., 2019; Lee et al., 2019; Li and Fan, 2019), in which biomarkers collected at each time go through a node of the RNN, and the output of the network in each later node is the prediction score. In the recent TADPOLE challenge (Azvan et al., 2020), the best performing team overall (ventricle volume, diagnosis, and cognitive score prediction) uses XGboost method (Chen and Guestrin, 2016); the best performing team in predicting ventricle volume alone uses data-driven disease progression model and machine learning (linear mixed effect model) (Venkatraghavan et al., 2018). Besides, Generative Adversarial Networks (GAN) have been applied to generate future images with or without AD pathology on the whole brain or in the MTL region (Bowles et al., 2018; Ravi et al., 2019).

To our knowledge, none of the DL longitudinal MRI analysis methods employed deep learning specifically as a means to derive a more effective disease progression and treatment evaluation biomarker for clinical trials. DL-based registration methods in which deformation fields are generated by a convolutional neural network (CNN) are an area of active research (Balakrishnan et al., 2018; Tustison et al., 2019; Yang et al., 2017). However, the impact of these methods on disease progression biomarkers in AD has not yet been evaluated.

## 4.2 Mathematical Intuition for the DeepAtrophy Algorithm

To better understand the constraints imposed on the output of the DeepAtrophy algorithm by the STO and RISI losses, we present a simplified mathematical analysis. Let $k$ denote an individual subject, let $\boldsymbol{T}^k = [T_0^k, T_1^k]$ be a time interval during which this subject is observed, and let $f^k(t): \boldsymbol{T}^k \to \mathbb{R}$ describe some scalar measure of disease burden in this individual (e.g., total amount of gray matter lost). Furthermore, let $F^k(s, t) = f^k(t) - f^k(s)$ describe the change in this measure of disease progression from time $s$ to time $t$.

If we assume that $f^k(t)$ is a monotonically increasing function of $t$ on $\boldsymbol{T}^k$, then it holds that

$$\text{sign } F^k(s,t) = \text{sign}[f^k(t) - f^k(s)] = \text{sign}(t-s) \quad \forall s,t \in \boldsymbol{T}^k. \tag{4}$$

Conversely, a function $f^k(t)$ that satisfies the condition above is monotonically increasing in t.

If we further assume that $f^k(t)$ is linear in $t$ on $\boldsymbol{T}^k$, i.e.,

$$f^k(t) = a_k t + b_k \quad \forall t \in \boldsymbol{T}^k,$$

and hence

$$F^k(s,t) = a_k(t-s) \quad \forall s,t \in \boldsymbol{T}^k,$$

then the following condition holds:

$$\frac{|F^k(s,t)|}{|F^k(s,u)|} = \frac{|t-s|}{|u-s|} \quad \forall s,t,u \in \boldsymbol{T}^k, \tag{5}$$

and conversely, when the above condition is satisfied, $f^k(t)$ is linear in $t$ on $\boldsymbol{T}^k$.

Let us consider the basic subnetwork in DeepAtrophy as a function that takes as input two images, $I^k(s)$ and $I^k(t)$, and that produces as its output the value $\tilde{F}_\theta\big(I^k(s), I^k(t)\big)$ that is an approximation of $F^k(s,t)$, i.e., a measure of change in disease progression between times $s$ and $t$. Here $\theta$ denotes the weights of the neural network learned during training. Conceptually, we can think of the basic subnetwork implicitly estimating the disease progression function $\tilde{f}_\theta\big(I^k(s)\big)$ for each of its inputs, and outputting the estimated change measure $\tilde{F}_\theta\big(I^k(s), I^k(t)\big) = \tilde{f}_\theta\big(I^k(t)\big) - \tilde{f}_\theta\big(I^k(s)\big)$. Then the STO loss, which is equivalent to (4), can be thought of imposing monotonicity on $\tilde{f}_\theta$ and the RISI loss, which is equivalent to (5), imposing linearity on $\tilde{f}_\theta$. We emphasize that $\widetilde{f}_\theta$ is not explicitly modeled or estimated in DeepAtrophy, and we think of it in purely implicit terms.

In practice, there are a few deviations in the actual implementation of DeepAtrophy and this analysis. The RISI loss is categorical rather than regression-based. Instead of the strict linearity condition (5), it imposes weaker constraints:

$$\frac{|F^k(s,t)|}{|F^k(s,u)|} \in \Gamma^p \quad \text{iff} \quad \frac{|t-s|}{|u-s|} \in \Gamma^p \quad \forall s,t,u \in \boldsymbol{T}^k; \ p = 0,\dots,P \qquad (6)$$

where $\Gamma^p \subset \mathbb{R}$ represent non-overlapping regions of the real line, i.e., $[1,2)$ or $[2,\infty)$. Furthermore, not all timepoints in $\boldsymbol{T}^k$ are considered, but only the ones for which scans are available. Also, the basic subnetwork has not one, but five output values, two of which enter into a cross-entropy STO loss (and can be thought of as corresponding to $\widetilde{F}_\theta$) and the other three provide additional input to the RISI loss. So while we cannot say that the RISI loss imposes linearity on the underlying measure of disease progression $\widetilde{f}_\theta$, the overall effect is for the network output $\widetilde{F}_\theta$ to have a larger magnitude over longer time intervals than over shorter intervals, with at least an approximately linear relationship. Given that, biologically, atrophy in neurodegenerative disease is not expected to be linear over long time intervals the relaxation of the linearity condition can be considered a strength of the DeepAtrophy framework.

## 4.3 DeepAtrophy Infers Temporal Information from Longitudinal Scan Pairs with Greater Accuracy than ALOHA Atrophy Measures

In Section 3.1 we compared the accuracy of DeepAtrophy and ALOHA in the tasks of inferring the temporal order of two scans (STO accuracy) and inferring which pair of scans has a longer interscan interval (RISI accuracy). Together, we refer to these two measures of accuracy as "temporal inference accuracy". DeepAtrophy indeed infers the true chronological order of pairs of same-subject scans presented to the network in arbitrary chronological order with excellent accuracy: 87.6%, 87.7%, 89.8%, and 93.2% for the control, preclinical AD, early prodromal AD, and late prodromal AD groups, respectively (Table 2, Figure 2). DeepAtrophy is also highly accurate (86.1% overall) at inferring which pair of scans from the same subject has the longer inter-scan interval. Both results represent a substantial improvement over ALOHA on the same test set. On the one hand, this is not surprising, since DeepAtrophy is explicitly formulated to maximize this accuracy through the STO and RISI losses, while ALOHA is a conventional hippocampal DBM method that uses deformable registration to measure change in hippocampal volume. On the other hand, it is somewhat surprising just how accurate DeepAtrophy is, particularly for the control and preclinical AD groups, at teasing out information about temporal order and time between scans just from image content.

We did not compare DeepAtrophy with other conventional techniques, but ALOHA is considered a state-of-the-art method (Das et al., 2012; L. Xie et al., 2020) and performs on par with other leading DBM (M. Lorenzi et al., 2015) and BSI techniques (Freeborough and Fox, 1997; Leung et al., 2010). By warping the baseline and follow-up images to halfway space in the registration and using a mesh to compute the volume change from the transformation field, ALOHA takes special care to reduce bias that may occur if baseline and follow-up scans are processed differently (Hua et al., 2016; Yushkevich et al., 2009). While we did not detect significant bias in ALOHA in our experiments, the variance was high, with only 77% of the pairwise atrophy measurements corresponding to hippocampal volume decrease. In contrast, the DeepAtrophy approach can correctly infer the scan order in roughly 90% of scans. For both methods, temporal inference accuracy generally increases with disease severity, which is to be expected, since the contribution of biological factors to changes in MRI image content is greater in more affected groups which experience higher rates of neurodegeneration. Increased sensitivity of the longitudinal measure and lower frequency of positive atrophy values (i.e., reports of hippocampal volume increase) in more affected groups is consistent with other atrophy measurement methods (Hua et al., 2016; Leung et al., 2010; Yushkevich et al., 2009).

An interesting finding of our study (see Supplementary S3) is that training a CNN only to infer temporal scan order (STO loss with very small weight of the RISI loss) is not enough to achieve a reliable AD progression biomarker. A CNN with a high relative weight of the STO loss does very well at detecting the presence of disease progression, but scales poorly with the magnitude of disease progression. This is revealed by the sublinear trajectories in the spaghetti plots in Figure 3-(a) and Supplemental Figure S1. Increasing the weight of the RISI loss makes the trajectories more linear, sensitizing the method to the magnitude of disease progression, at a moderate cost for short-term detection of progression (more incorrect temporal order predictions). The differences between groups are also more pronounced with a higher weight of the RISI loss. The ability of a biomarker to quantify the magnitude and not just presence of progression is important because in a clinical trial both the treatment and the placebo cohort are expected to have disease progression, and the role of a biomarker is to detect a subtle difference in rates of progression. However, it is worth noting even when using a RISI loss with a high weight, the trajectories remain sublinear, i.e., the PAIIR measure remains somewhat more sensitive to the presence of progression than to its magnitude. In this sense, the network prediction has a lower transitivity and cannot completely be a surrogate to volume change measurement, which appears to follow a linear trajectory. We empirically chose 1 as the weight of the RISI loss, and the experiments in Supplementary S3 were performed as a *post hoc* analysis and did not influence our weighting of the STO and RISI losses.

One critical question is whether the higher temporal inference accuracy in DeepAtrophy reflects greater sensitivity to disease progression, or whether other non-biological factors that are not independent of time, i.e., systematic, are present. For example, in a single-site longitudinal study, a change in scanner hardware or protocol parameters at certain points over the duration of the study would result in differences in image content that are systematic yet not biological (e.g., scans acquired later in the study might have greater white/gray tissue contrast). A CNN would easily detect this difference resulting in high temporal inference accuracy. However, in such a scenario, we would expect the STO accuracy of the CNN to be high, but less so the RISI accuracy. Most importantly, if such a CNN was primarily detecting factors that are systematic but non-biological, we would not expect to observe significant differences in CNN output between less affected and more affected individuals. The fact that DeepAtrophy has high RISI accuracy, does slightly better than ALOHA at group separation (particularly preclinical AD vs. control groups), and is trained on a multi-site multi-scanner dataset, makes it very unlikely that non-biological factors are driving temporal inference accuracy. In future work, it would be informative to relate data on software and hardware changes at ADNI sites during the ADNI2/GO phases the study to DeepAtrophy measures, and thus determine to what extent these measures are impacted by these systematic but non-biological changes.

Furthermore, in the Supplementary S2, we tested DeepAtrophy on nine subjects scanned on the same day. The STO accuracy in this experiment was reduced to around 50%, which means that the network could not infer the temporal order of two scans in a short time interval much better than chance. This back-to-back experiment essentially rules out the influence of additional non-systematic factors, such as inconsistent preprocessing of scan pairs.

## 4.4 Potential of DeepAtrophy as a Biomarker of Therapeutic Efficacy

A key result in Section 3.3 is that using longitudinal MRI scans from the first year of each ADNI participant's enrollment (180-400 days window for follow-up MRIs) both DeepAtrophy and ALOHA were able to detect significant differences between the control group and the early/late prodromal AD groups, while only DeepAtrophy detected a significant difference between the control and preclinical AD groups. Overall, the effect sizes in group comparisons were larger, in absolute terms, for DeepAtrophy than ALOHA, suggesting greater sensitivity to differences in rates of disease progression. This was also the case when we considered a 400-800-day follow-up window.

Studies commonly evaluate longitudinal biomarkers in AD by estimating the sample size needed to power a hypothetical clinical trial in which the experimental treatment is expected to reduce the rate of disease progression by 25% relative to the healthy aging (Holland et al., 2012a; Hua et al., 2016; Pegueroles et al., 2017; Yushkevich et al., 2009). Sample size estimates reported in the literature for longitudinal MRI-based biomarkers are generally smaller than for cognitive testing (Ard and Edland, 2011; Cullen et al., 2014; Weiner et al., 2015; Xie et al., 2020). Most sample size estimates reported in the literature involve hypothetical clinical trials in MCI or AD. In the MIRIAD challenge (Cash et al., 2015), the smallest reported sample sizes for a hypothetic 12 month clinical trial in AD were 190 (95% CI: 146 to 268) and 158 (95% CI: 116 to 228) for left and right hippocampi, respectively. The dataset in this challenge was collected on the same scanner by the same physician for a long period of time and controlled for age. In the original report of ALOHA (Das et al., 2012), a sample size of 269 (based on a one-sided test, corresponds to 343 for two-sided) was estimated for a hypothetical one-year trial in MCI (regardless of beta-amyloid status) using a hippocampal volume atrophy measure derived from longitudinal high-resolution T2-weighted MRI; and sample size of 325 (414 two-sided) when using T1-weighted MRI. In subsequent a comparison of FreeSurfer (FS), Quarc, and KN-BSI T1-MRI analysis methods in Holland et al. (2012), the minimum sample size reported for a one-year trial in late MCI was 327 (95% CI: 209 to 585). However, even though the sample size in these studies was reported for a one-year trial, the annualized atrophy rates used to estimate these sample sizes used *longitudinal scans with up to three years follow-up*. By contrast, in a hypothetical one-year clinical trial in late MCI, the sample size using DeepAtrophy is estimated to be only 253 (95% CI: 136 to 503), and unlike the above studies, this estimate is based on one-year follow-up data. The corresponding estimate for ALOHA is 455 (95% CI: 209 to 1195), consistent with the other DBM studies above. This suggests that DeepAtrophy improves on the state-of-the-art conventional methods for disease progression quantification in the context of symptomatic AD.

Compared to MCI/AD, there has been relatively less work on estimating the sample size needed to power a hypothetical clinical trial in preclinical AD. Insel et al. (2019) report a sample size of 2000 for a 4-year clinical trial using PACC as the outcome measure. Holland et al. (2012b) performed sample size estimation for a three-year clinical trial in preclinical AD, where they reported n = 1763 (95% CI: [400, >100000]) needed to detect a 25% reduction in longitudinal hippocampus change rate relative to controls, applied to data collected in 3 years. Similarly, the sample size estimated for a hypothetical 3-year clinical trial for a 25% reduction in hippocampus volume change by Bertens et al. (2017) is 279 (95% CI: [197, 426]). Xie et al. (2020) reported the results of the ALOHA analysis described in the current study and reported sample sizes consistent with the results in Table 4. For a one-year trial with a 25% reduction in disease progression in the preclinical AD group relative to the control group, DeepAtrophy requires the sample size of 4492 (95% CI: [733, >100000]), compared to 6832 (95% CI: [1253, >100000]) for ALOHA. While these are very large numbers, to our knowledge, ours is the first study in the literature to report sample size for preclinical AD within such a short trial. For a more realistic two-year trial duration, the sample size estimates are 1499 (95% CI: [403, 28998]) for DeepAtrophy and 2328 (95% CI: [636, 120045]) for ALOHA. If we assume that hypothetical intervention will reduce the rate of atrophy in preclinical AD by 50% rather than 25% relative to healthy aging (i.e., from ~0.8% to ~0.65% a year), the sample size to power a clinical trial is a much more practical 374 (95% CI: [101, 6103]) for DeepAtrophy and 582 (95% CI: [155, 26301]) for ALOHA. Overall, DeepAtrophy gives reasonable and competitive estimates for powering a two-year clinical trial of preclinical AD. This is particularly impressive because no subjects with preclinical AD were included in the training set.

Although the 95% confidence intervals for the DeepAtrophy and ALOHA sample size estimates overlap, and we cannot statistically conclude that the former is better at powering a clinical trial, in absolute terms, the DeepAtrophy sample sizes are smaller in all scenarios in Table 4, and also appear to improve on published sample sizes for MRI-based and cognitive measures. Moreover, as we discuss below, the current DeepAtrophy is the first attempt to leverage temporal order inference in a disease progression quantification algorithm, hence it leaves room for future improvement.

## 4.5 Limitations and Future Work

Perhaps the main limitation of DeepAtrophy compared to DBM/BSI techniques is that it provides a holistic interpretation of change over time in a longitudinal scan pair and does not shed light on neurodegeneration in specific anatomical regions. Whereas ALOHA can provide measures of change in specific anatomical regions (hippocampus, Brodmann area 35), DeepAtrophy yields only a single measure for the MTL ROI, measured in terms of mismatch between inferred and actual interscan interval (PAIIR). This limits the interpretability of the DeepAtrophy results, which is a common limitation of many deep learning image analysis approaches. However, existing approaches for interpretation of deep learning models (e.g., attention mapping, gradient-based techniques (Selvaraju et al., 2016; Zhang et al., 2018) or weakly supervised learning (Durand et al., 2017)) can be readily applied to DeepAtrophy. A more ambitious approach would be to combine DeepAtrophy and DBM/ALOHA in a single end-to-end deep learning algorithm that would integrate deformable registration and generation of Jacobian determinant maps into the network (similar to (Balakrishnan et al., 2018; Tustison et al., 2019; Yang et al., 2017) ), and would be trained to maximize STO and RISI losses, perhaps along with standard image similarity metrics.

We hypothesize that such an algorithm would have the improved temporal inference accuracy of DeepAtrophy, while also yielding accurate maps of tissue expansion and compression.

Another limitation of our approach is that it focuses on pairs of scans. When three or more scans are available, we use linear models to infer a summary PAIIR measure from pairwise PAIIR data. Directly incorporating multiple scans into the network, perhaps in a recurrent neural network architecture may offer additional efficiencies and improved accuracy over the current approach.

As noted above, the RISI and STO losses offer tradeoffs between sensitivity to subtle change and sensitivity to magnitude of disease progression. Our experiments were performed with a very small validation set and so we settled on an empirical weighting of the losses. A more elegant strategy would have been to optimize the weight of the loss on the validation set, e.g., by choosing the weight that leads to best discrimination between preclinical AD and control groups. However, given the data sparsity for preclinical AD, we opted to put all the available participants with preclinical AD into the held-out test set. With additional preclinical AD longitudinal datasets, such as the A4 study (Sperling et al., 2014), such analysis may be possible. A related potential point of improvement is the categorical, and somewhat arbitrary nature, of the RISI loss. Further research, including the modification of the underlying image classification deep network (Xie et al., 2020), may lead to better trainability of a regression-type RISI loss, in turn leading to a more linear relationship of predicted interscan interval with actual interscan interval (Figure 3-(a)) and perhaps greater sensitivity to disease progression.

## 5. Conclusion

In this paper, we showed that a deep learning network, DeepAtrophy, can infer the temporal order of same-subject longitudinal MRI scans, as well as deduce which pair of same-subject scans has a longer interscan interval, with excellent accuracy. The design of DeepAtrophy encapsulates the underlying assumption that in the context of Alzheimer's disease, image changes that are systematic with time are related to neurodegeneration, while image changes that are independent of time are related to non-biological factors such as patient motion and MRI artifact. We used the output layer of this network to construct a measure of disease progression, formulated as the mismatch between the actual interscan interval for a pair of MRI scans, and the interscan interval predicted by our network. Our experiments on a held-out test set show that this mismatch measure is sensitive to differences in rates of disease progression between the earliest detectable stage of Alzheimer's disease and healthy aging, suggesting that it may have advantages over conventional deformation-based longitudinal MRI biomarkers that currently serve as outcome measures in clinical trials of disease-modifying treatments in Alzheimer's disease.

# Acknowledgement

This work was supported by National Institute of Health (NIH) (grant numbers R01-AG056014, R01-AG040271, P30-AG010124, R01-EB017255, R01-AG055005).

Data collection and sharing for this project was funded by the Alzheimer's Disease Neuroimaging Initiative (ADNI) (National Institutes of Health Grant U01 AG024904) and DOD ADNI (Department of Defense award number W81XWH-12-2-0012). ADNI is funded by the National Institute on Aging, the National Institute of Biomedical Imaging and Bioengineering, and through generous contributions from the following: AbbVie, Alzheimer's Association; Alzheimer's Drug Discovery Foundation; Araclon Biotech; BioClinica, Inc.; Biogen; Bristol-Myers Squibb Company; CereSpir, Inc.; Cogstate; Eisai Inc.; Elan Pharmaceuticals, Inc.; Eli Lilly and Company; EuroImmun; F. Hoffmann-La Roche Ltd and its affiliated company Genentech, Inc.; Fujirebio; GE Healthcare; IXICO Ltd.; Janssen Alzheimer Immunotherapy Research & Development, LLC.; Johnson & Johnson Pharmaceutical Research & Development LLC.; Lumosity; Lundbeck; Merck & Co., Inc.; Meso Scale Diagnostics, LLC.; NeuroRx Research; Neurotrack Technologies; Novartis Pharmaceuticals Corporation; Pfizer Inc.; Piramal Imaging; Servier; Takeda Pharmaceutical Company; and Transition Therapeutics. The Canadian Institutes of Health Research is providing funds to support ADNI clinical sites in Canada. Private sector contributions are facilitated by the Foundation for the National Institutes of Health (www.fnih.org). The grantee organization is the Northern California Institute for Research and Education, and the study is coordinated by the Alzheimer's Therapeutic Research Institute at the University of Southern

California. ADNI data are disseminated by the Laboratory for Neuro Imaging at the University of Southern California.

# Supplementary Material

## S1. Characteristics of the ADNI2/GO Participants with 180-to-400 Days Scans and 400-to-800 Days Scans

When evaluating the results for scans only in a short interval in Section 3.3 and 3.4, individuals who only have scans out of the range (180 to 400 days or 400 to 800 days) was not included. The number of individuals in the test set were reduced from 326 to 256 (for scans within 180 to 400 days) and 246 (for scans within 400 to 800 days) respectively. In Supplemental Table S1 the age, sex, education year, and MMSE of the subjects in both intervals are summarized.

Supplemental Table S1. Characteristics of the held-out test dataset for participants whose T1 MRI scans were used for the DeepAtrophy, ALOHA, and PACC experiments for group analysis and sample size calculation in Figure 4, Table 4, and Supplemental Figure S2. Only scans within 180 to 400 days interval (left) and 400 to 800 days interval (right) are included in the experiment. Abbreviations: ALOHA = Automatic Longitudinal Hippocampal Atrophy software/package; PACC = Preclinical Alzheimer's Cognitive Composite; n = number of subjects; A+/A- = beta-amyloid positive/negative; NC = cognitively normal adults; eMCI = early mild cognitive impairment; lMCI = late mild cognitive impair; Edu = years of education; MMSE = mini-mental state examination; F = female; M = male.

| | Test Set (180 to 400 days scans, n = 256) | | | | Test Set (400 to 800 days scans, n = 246) | | | |
|---|---|---|---|---|---|---|---|---|
| | A- NC (n = 75) | A+ NC (n = 58) | A+ eMCI (n = 68) | A+ lMCI (n = 55) | A- NC (n = 78) | A+ NC (n = 67) | A+ eMCI (n = 61) | A+ lMCI (n = 40) |
| Age (years) | 72.5 (6.5) | 75.8 (5.8)** | 74.0 (7.0) | 72.4 (6.7) | 72.1 (6.4) | 75.5 (5.4)*** | 73.3 (6.8) | 71.7 (6.5) |
| Sex | 40F 35M | 35F 23M | 34F 34M | 26F 29M | 43F 35M | 44F 23M | 31F 30M | 19F 21M |

| Edu (years) | 16.8 (2.5) | 16.1 (2.6) | 15.2 (3.0)*** | 16.5 (2.7) | 16.5 (2.4) | 15.9 (2.9) | 15.1 (2.8)** | 16.4 (2.7) |
|---|---|---|---|---|---|---|---|---|
| MMSE | 29.2 (1.2) | 29.0 (1.2) | 28.1 (1.6)**** | 26.7 (2.0)**** | 29.0 (1.3) | 29.0 (1.1) | 28.0 (1.6)*** | 27.1 (2.0)**** |

*Notes*: *, $p < 0.05$; **, $p < 0.01$; ***, $p < 0.001$; ****, $p < 0.0001$. All statistics in the train and test set are in comparison to the corresponding A- NC group. *Standard deviation* is reported in parenthesis. Independent two-sample *t*-test (continuous variables with normal distribution, for age and education), Mann–Whitney *U* test (continuous variable with non-normal distribution, for MMSE) and contingency $\chi^2$ test (sex) were performed.

## S2. Temporal Order Inference Accuracy of DeepAtrophy on a Held-out Test-retest MRI Dataset

To confirm that the high temporal inference accuracy of DeepAtrophy in ADNI2/GO longitudinal scans is not driven by systematic factors that are non-biological (e.g., inconsistent preprocessing of scan pairs, programming or scripting error), we tested our network on a held-out dataset of 9 subjects scanned twice on the same day. These T1-weighted MRI scans were collected at University of California, San Francisco (UCSF) and went through the same preprocessing steps as described in Section 2.1 (Das et al., 2012). This generates 36 pairs of test samples since we have scans for the left and right hemispheres and we can input pairs to the network in the correct or reversed order. For subjects scanned with such short time interval, no atrophy is expected. After inputting each scan pair to the basic sub-network in Figure 1, we observed that 14 of the 36 scans are correctly temporally ordered by the network's output layer, and that 16 out of 36 scans are predicted with a positive label (noting that if the basic sub-network predict positive for all pairs, then the accuracy would also be 50%). A binomial test shows that neither accuracy is significantly different from chance (p-value = 0.243, 0.618, respectively). This suggests that progressive changes detected by DeepAtrophy in longitudinal data from ADNI are related to biological factors.

## S3. Post Hoc Analysis of the Weight of the Relative Interscan Interval (RISI) Loss

A spaghetti plot to compare the influence of different weights of the RISI loss in DeepAtrophy is shown in supplemental Figure S1. The weights are 0, 0.1, 1, and 10 respectively from (a) to (d), while the weight of the STO loss is fixed at 1. With a lower RISI loss, subplot S1 (a) has a higher accuracy in correct temporal order prediction (most of the predictions are above 0), but almost no difference in the magnitude of the PII measure between 3, 6 and 12-month interscan intervals. Subplot S1 (d), however, shows a PII measure that increases more with respect to time, but also makes more errors for temporal order prediction, especially in the A- NC group. Subplot S1 (d) also has the best group differentiation in all groups. In the main text, we report the results using the empirically chosen weight of 1 for the RISI loss.

Supplemental Figure S1. The effect of adjusting the weight of the RISI loss relative to the STO loss in DeepAtrophy training on the PII disease progression measurement. The RISI loss weights are 0, 0.1, 1, and 10 respectively from (a) to (d). With zero RISI loss weight (a), DeepAtrophy achieves higher accuracy in temporal order prediction (most of the PII values are above zero), but the trajectories are nearly constant across the 3, 6 and 12-month time points. With higher RISI loss weight (d), the trajectories are more linear over time but there are slightly more errors for temporal order prediction, especially in the A- NC group. Abbreviations: RISI = relative inter-scan interval; STO = scan temporal order; PII = predicted interscan interval; A+/A- = beta-amyloid positive/negative; NC = cognitively normal adults; eMCI = early mild cognitive impairment; lMCI = late mild cognitive impair.

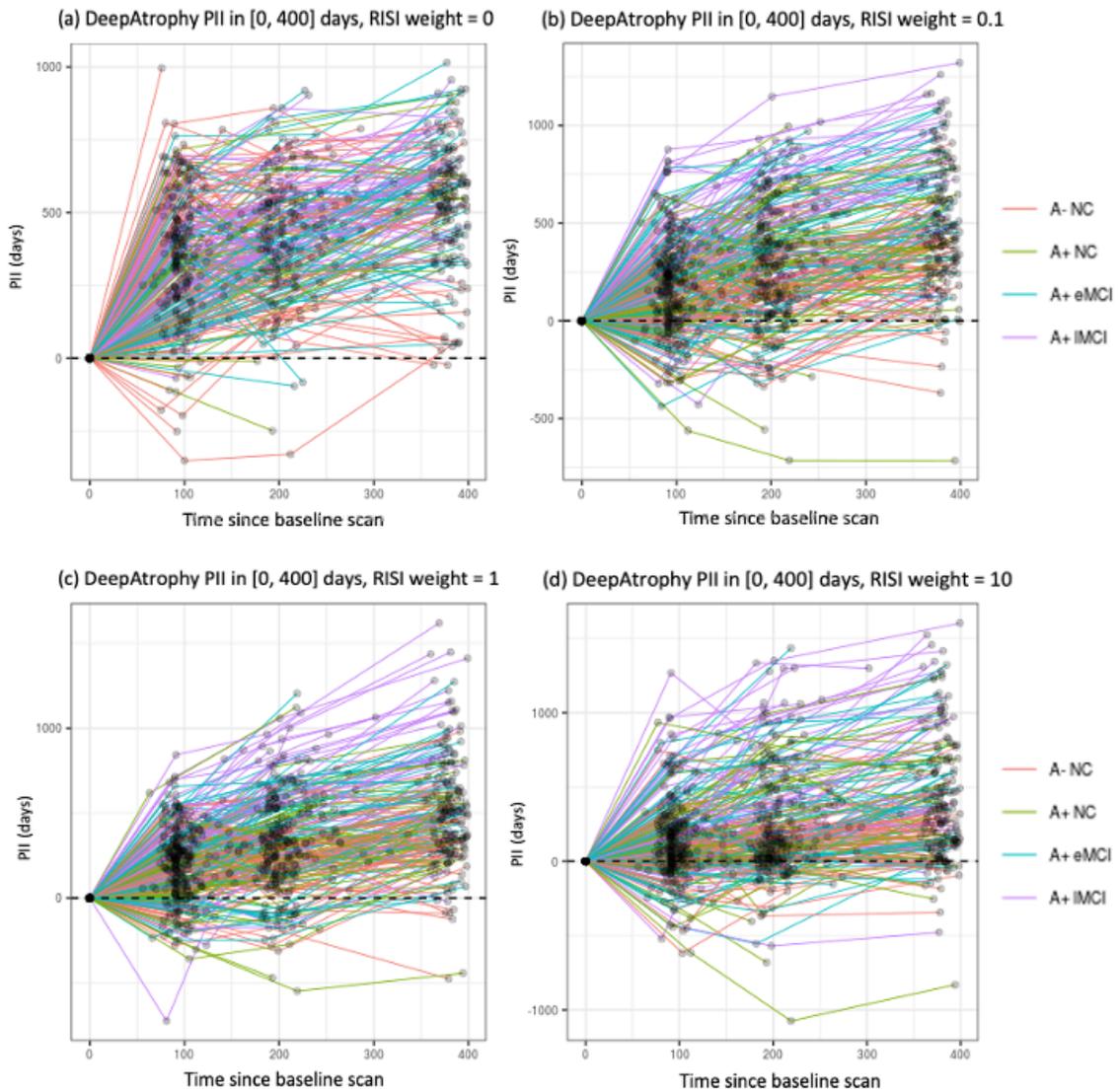

## S4. Group Differences Detected by DeepAtrophy and ALOHA in Data with 400 to 800 Day Follow-Up Range

Supplemental Figure S2 gives a plot corresponding to Figure 4 but using follow-up longitudinal scans between 400 and 800 days from baseline. The plot compares the sensitivity of DeepAtrophy, ALOHA and PACC to differences in rates of disease progression between the four diagnostic groups (A- NC vs. A+ NC, A+ eMCI, and A+ lMCI). All other processing steps are the same as in Section 3.3 (Figure 4).

Supplemental Figure S2. Comparison of the ability of DeepAtrophy PAIIR measure, ALOHA atrophy rate, and PACC score change rate to detect differences in the rates of disease progression between beta-amyloid-negative normal controls (A- NC) and three disease groups: beta-amyloid-positive preclinical AD (A+ NC), early prodromal AD (A+ eMCI) and late prodromal AD (A+ lMCI) using follow-up timepoints between 400 to 800 days from baseline. In each subplot, independent samples t-test were conducted, after correcting for age, in comparison with the control group and the p-values were shown for each comparison. With the same subjects and pairs, DeepAtrophy yields higher level of significance in differentiating all disease groups from A- NC compared to ALOHA and PACC, suggesting that DeepAtrophy may be a better biomarker to detect disease progression at an early disease stage. Abbreviations: PAIIR = predicted-to-actual interscan interval ratio; ALOHA = Automatic Longitudinal Hippocampal Atrophy software/package; PACC = Preclinical Alzheimer's Cognitive Composite; A+/A- = beta-amyloid positive/negative; NC = cognitively normal adults; eMCI = early mild cognitive impairment; lMCI = late mild cognitive impair; N = number of subjects in the diagnosis group.

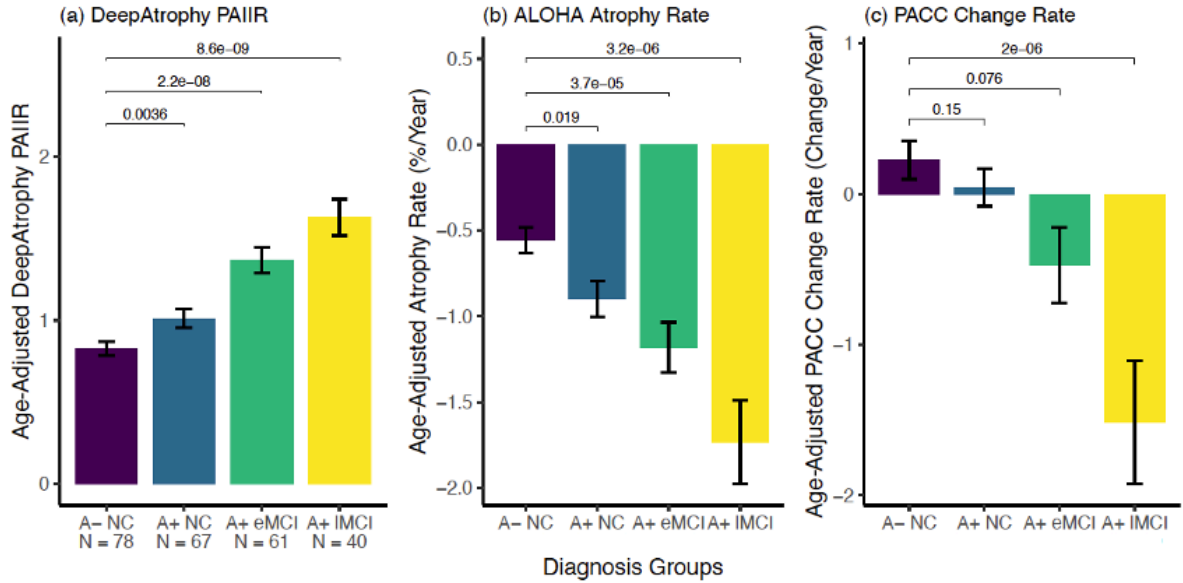